\documentclass[11pt,a4paper]{article}
\usepackage[hyperref]{emnlp2020}
\usepackage{times}
\usepackage{latexsym}

\usepackage{microtype}

\aclfinalcopy 
\def\aclpaperid{3464} 

\usepackage{microtype}
\usepackage{latexsym} 
\usepackage{graphicx}
\usepackage{wrapfig}
\usepackage{amssymb}
\usepackage{color,xcolor,colortbl}
\usepackage{enumitem}
\usepackage{soul}
\usepackage{amsmath}
\usepackage{url}
\usepackage{algorithm}
\usepackage{algorithmic}
\usepackage{booktabs}
\usepackage{bm,bbm}
\usepackage{mathtools}
\usepackage{array}
\usepackage{multirow}
\usepackage{subcaption}
\usepackage{pbox}
\usepackage{pifont}
\usepackage{arydshln}
\usepackage{dblfloatfix}
\usepackage{relsize}

\newcommand\numberthis{\addtocounter{equation}{1}\tag{\theequation}}

\definecolor{darkblue}{rgb}{0.0, 0.0, 0.55}
\setul{0.5ex}{0.3ex} 
\setulcolor{blue} 
\usepackage[T1]{fontenc}
\usepackage[scaled=.8]{beramono}

\title{Learning to Fuse Sentences with Transformers for Summarization}

\author{Logan Lebanoff$^\dagger$ \,\, Franck Dernoncourt$^\S$\\
\textbf{Doo Soon Kim$^\S$ \,\, Lidan Wang$^\S$ \,\, Walter Chang$^\S$ \,\, Fei Liu$^\dagger$}
\\[0.8em]
$^\dagger$University of Central Florida \quad $^\S$Adobe Research\\[0.8em]
\texttt{loganlebanoff@knights.ucf.edu} \quad \texttt{feiliu@cs.ucf.edu}\\ 
\texttt{\{dernonco,dkim,lidwang,wachang\}@adobe.com}
}

\begin{document}

\maketitle

\begin{abstract}

The ability to fuse sentences is highly attractive for summarization systems because it is an essential step to produce succinct abstracts.
However, to date, summarizers can fail on fusing sentences.
They tend to produce few summary sentences by fusion or generate incorrect fusions that lead the summary to fail to retain the original meaning. 
In this paper, we explore the ability of Transformers to fuse sentences and propose novel algorithms to enhance their ability to perform sentence fusion by leveraging the knowledge of \emph{points of correspondence} between sentences.
Through extensive experiments, we investigate the effects of different design choices on Transformer's performance.
Our findings highlight the importance of modeling points of correspondence between sentences for effective sentence fusion. 

\end{abstract}

\section{Introduction}
\label{sec:intro}

A renewed emphasis must be placed on sentence fusion in the context of neural abstractive summarization.
A majority of the systems are trained end-to-end~\cite{see-etal-2017-get,paulus2018a,narayan-etal-2018-dont,chen-bansal-2018-fast,gehrmann-etal-2018-bottom,liu-lapata-2019-hierarchical}, where an abstractive summarizer is rewarded for generating summaries that contain the same words as human abstracts, measured by automatic metrics such as ROUGE~\cite{lin-2004-rouge}.
A summarizer, however, is not rewarded for correctly fusing sentences.
In fact, when examined more closely, only few sentences in system abstracts are generated by fusion~\cite{falke-etal-2019-ranking,lebanoff-etal-2019-analyzing}.
For instance, 6\% of summary sentences generated by Pointer-Gen~\cite{see-etal-2017-get} are through fusion, whereas human abstracts contain 32\% fusion sentences.
Moreover, sentences generated by fusion are prone to errors.
They can be ungrammatical, nonsensical, or otherwise ill-formed.
There is thus an urgent need to develop neural abstractive summarizers to fuse sentences properly.

\begin{figure*}
\centering
\includegraphics[width=6in]{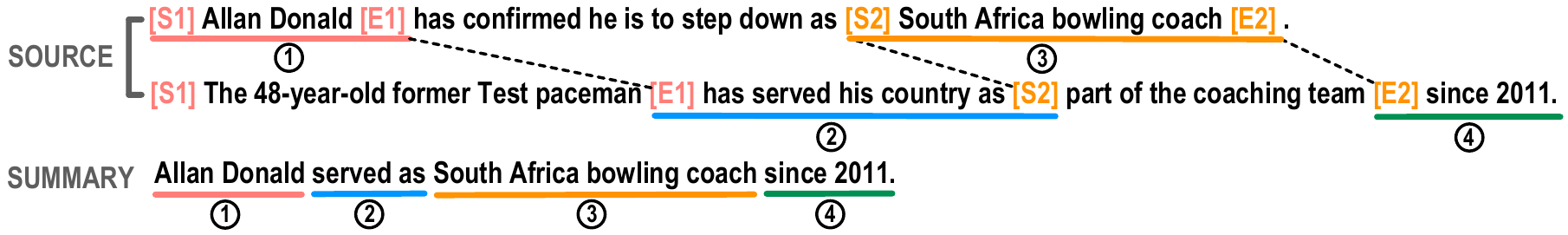}
\vspace{-0.05in}
\caption{
Sentence fusion involves determining what content from each sentence to retain, and how best to weave text pieces together into a well-formed sentence.
Points of correspondence (PoC) are text chunks that convey the same or similar meanings, e.g., \emph{Allan Donald} and \emph{The 48-year-old former Test paceman}, \emph{South Africa bowling coach} and \emph{part of the coaching team}. 
}
\label{fig:example_fusion}
\vspace{-0.1in}
\end{figure*}

The importance of sentence fusion has long been recognized by the community before the era of neural text summarization.
The pioneering work of Barzilay et al.~\shortcite{barzilay-etal-1999-information} introduces an information fusion algorithm that combines similar elements across related text to generate a succinct summary.
Later work, such as~\cite{marsi-krahmer-2005-explorations,filippova-strube-2008-sentence,elsner-santhanam-2011-learning,thadani-mckeown-2013-supervised,mehdad-etal-2013-abstractive}, builds a dependency or word graph by combining syntactic trees of similar sentences, then employs integer linear programming to decode a summary sentence from the graph. 
Most of these studies have assumed a set of \emph{similar} sentences as input, where fusion is necessary to reduce repetition.
Nonetheless, humans do not limit themselves to combine similar sentences.
In this paper, we pay particular attention to fuse \emph{disparate} sentences that contain fundamentally different content but remain related to make fusion sensible~\cite{elsner-santhanam-2011-learning}.
In Figure~\ref{fig:example_fusion}, we provide an example of a sentence fusion instance.

We address the challenge of fusing disparate sentences by enhancing the Transformer architecture~\cite{NIPS2017_7181} with \emph{points of correspondence} between sentences, which are devices that tie two sentences together into a coherent text.
The task of sentence fusion involves choosing content from each sentence and weaving the content pieces together into an output sentence that is linguistically plausible and semantically truthful to the original input.
It is distinct from~\citet{geva-etal-2019-discofuse} that connect two sentences with discourse markers.
Our contributions are as follows.
\begin{itemize}[topsep=5pt,itemsep=0pt,leftmargin=*]
\item 
We make crucial use of \emph{points of correspondence} (PoC) between sentences for information fusion.
Our use of PoC was initiated by the current lack of understanding of how sentences are combined in neural text summarization.

\item 
We design new sentence fusion systems and experiment with a fusion dataset containing quality PoC annotations as the test bed for this investigation.
Our findings highlight the importance of modeling points of correspondence for fusion.\footnote{Our code is publicly available at {\url{https://github.com/ucfnlp/sent-fusion-transformers}}}

\end{itemize}

\begin{figure*}
\centering
\includegraphics[width=5in]{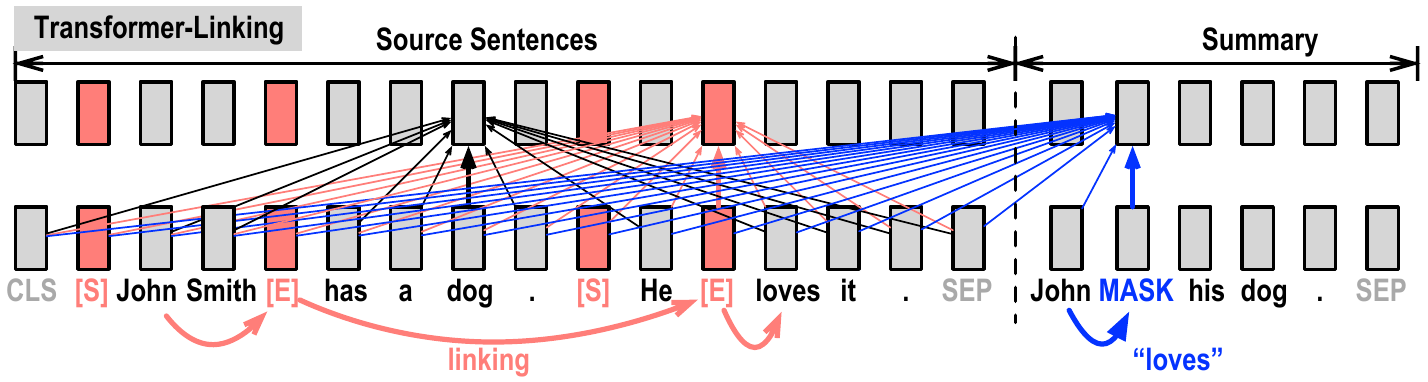}
\vspace{-0.05in}
\caption{
Our \textsc{Trans-Linking} model facilitates summary generation by reducing the shifting distance, allowing the model attention to shift from ``John'' to the tokens ``[E]'' then to ``loves'' for predicting the next summary word.
}
\label{fig:architecture}
\end{figure*}

\section{Method}
\label{sec:method}

A PoC is a pair of text chunks that express the same or similar meanings. 
In Fig.~\ref{fig:example_fusion}, \emph{Allan Donald} vs. \emph{The 48-year-old former Test paceman}, \emph{South Africa bowling coach} vs. \emph{part of the coaching team} are two PoCs.
The use of alternative expressions for conveying the same meanings is standard practice in writing, as it increases lexical variety and reduces redundancy.
However, existing summarizers cannot make effective use of these expressions to establish correspondence between sentences, often leading to ungrammatical and nonsensical outputs.

\subsection{Transformer with Linking}
\label{sec:linking}

It is advantageous for a Transformer model to make use of PoC information for sentence fusion.
While Transformer-based pretrained models have had considerable success~\cite{devlin-etal-2019-bert,NIPS2019_9464,lewis2019bart},
they primarily feature pairwise relationships between \emph{tokens}, but not PoC mentions, which are are \emph{text chunks} of varying size.
Only to a limited extent do these models embed knowledge of coreference~\cite{clark-etal-2019-bert}, and there is a growing need for incorporating PoC linkages explicitly in a Transformer model to enhance its ability to perform sentence fusion.

We propose to enrich Transformer's source sequence with \emph{markups} that indicate PoC linkages.
Here PoC information is assumed to be available for any fusion instance (details in \S\ref{sec:experiments}).
We introduce special tokens (\textsf{[S$_k$]} and \textsf{[E$_k$]}) to mark the start and end of each PoC mention; 
all mentions pertaining to the $k$-th PoC share the same start/end tokens.
An example is illustrated in Figure~\ref{fig:example_fusion}, where \emph{Allan Donald} and \emph{The 48-year-old former Test paceman} are enriched with the same special tokens.
We expect special tokens to assist in linking coreferring mentions, creating long-range dependencies between them and encouraging the model to use these mentions interchangeably in generation (Figure~\ref{fig:architecture}). 
The model is called ``\textsc{Trans-Linking}.''

Our Transformer takes as input a sequence $\mathcal{S}$ formed by concatenating the source and summary sequences.
Let $\mathbf{H}^l=[\mathbf{h}_1^l,\ldots,\mathbf{h}_{|\mathcal{S}|}^l]$ be hidden representations of the $l$-th layer of a decoder-only architecture.
An attention head transforms each vector respectively into query ($\mathbf{q}_i$), key ($\mathbf{k}_j$) and value ($\mathbf{v}_j$) vectors.
The attention weight $\alpha_{i,j}$ is computed for all pairs of tokens by taking the scaled dot product of query and key vectors and applying softmax over the output (Eq.~(\ref{eq:alpha})).
$\alpha_{i,j}$ indicates the importance of token $j$ to constructing $\mathbf{h}_i^{l}$ of the current token $i$.
\begin{align*}
&\alpha_{i,j} = \frac{\exp(\mathbf{q}_i^\top \mathbf{k}_j/\sqrt{d_k} + \mathcal{M}_{i,j})}{\sum_{j'=1}^{|\mathcal{S}|} \exp(\mathbf{q}_i^\top \mathbf{k}_{j'}/\sqrt{d_k} + \mathcal{M}_{i,j'})}
\numberthis\label{eq:alpha}
\end{align*}

We utilize a mask $\mathcal{M} \in \mathbb{R}^{|\mathcal{S}| \times |\mathcal{S}|}$ to control the attention of the model (Eq.~(\ref{eq:AttentionMask})).
$\mathcal{M}_{i,j}=0$ allows token $i$ to attend to $j$ and $\mathcal{M}_{i,j}=-\infty$ prevents $i$ from attending to $j$ as it leads $\alpha_{i,j}$ to be zero after softmax normalization.
Similar to~\cite{NIPS2019_9464},
a source token ($i \leq |\mathbf{x}|$) can attend to all other source tokens ($\mathcal{M}_{i,j} = 0$ for $j \leq |\mathbf{x}|$). 
A summary token ($i > |\mathbf{x}|$) can attend to all tokens including itself and those prior to it ($\mathcal{M}_{i,j} = 0$ for $j \leq i$).
The mask $\mathcal{M}$ provides desired flexibility in terms of building hidden representations for tokens in  $\mathcal{S}$.
The output of the attention head is a weighted sum of the value vectors $\mathbf{h}_i^l = \sum_{j=1}^{|\mathcal{S}|} \alpha_{i,j} \mathbf{v}_j$. 
\begin{equation}\label{eq:AttentionMask}
\mathcal{M}_{i,j} = \left\{
\begin{split}
&0 \quad\quad \text{if} \ j \leq \max(i, |\mathbf{x}|)\\
&-\infty \quad \text{otherwise}
\end{split}
\right.
\end{equation}

We fine-tune the model on a sentence fusion dataset (\S\ref{sec:experiments}) using a denoising objective, where 70\% of the summary tokens are randomly masked out.
The model is trained to predict the original tokens conditioned on hidden vectors of \textsf{\scriptsize MASK} tokens:
$\mathbf{o}=\mbox{softmax}(\mathbf{W}^O \mbox{GeLU}(\mathbf{W}^h \mathbf{h}_{\mbox{\scriptsize\textsf{MASK}}}^L)))$,
where parameters $\mathbf{W}^O$ are tied with token embeddings.
By inserting markup tokens, our model provides a soft linking mechanism to allow mentions of the same PoC to be used interchangeably in summary generation.
As shown in Figure~\ref{fig:architecture}, without PoC linking, the focus of the model attention has to shift a long distance from ``John'' to ``loves'' to generate the next summary word.
Their long-range dependency is not always effectively captured by the model.
In contrast, our \textsc{Trans-Linking} model substantially reduces the shifting distance, allowing the model to hop to the special token ``[E]'' then to ``loves,'' facilitating summary generation.

\subsection{Transformer with Shared Representation}
\label{sec:shared_attention}

We explore an alternative method to allow mentions of the same PoC to be connected with each other.
Particularly, we direct one attention head to focus on tokens belonging to the same PoC, allowing these tokens to share semantic representations, similar to Strubell et al.~\shortcite{strubell-etal-2018-linguistically}.
Sharing representation is meaningful as these mentions are related by complex morpho-syntactic, syntactic or semantic constraints~\cite{grosz-etal-1995-centering}.

Let $\mathbf{z}$=$\{z_1, \ldots, z_{|\mathbf{z}|}\}$ be a sequence containing PoC information, where
$z_i \in \{0,\ldots,\textsf{K}\}$ indicates the index of PoC to which the token $\mathbf{x}_i$ belongs.
$z_i$=$0$ indicates $\mathbf{x}_i$ is not associated with any PoC.
Our \textsc{Trans-ShareRepr} model selects an attention head $h$ from the $l$-th layer of the Transformer model.
The attention head $h$ governs tokens that belong to PoCs ($z_i \neq 0$). 
Its hidden representation $\mathbf{h}_i^l$ is computed by modeling only pairwise relationships between token $i$ and any token $j$ of the same PoC ($z_i=z_j$; Eq.~(\ref{eq:mask-share})), while other tokens are excluded from consideration. 
\begin{equation}\label{eq:mask-share}
\mathcal{M}_{i,j}^h = \left\{
\begin{split}
&0 \quad\quad \text{if} \ i,j \leq |\mathbf{x}|\ \&\ z_i = z_j\\
&-\infty \quad \text{otherwise}
\end{split}
\right.
\end{equation}

For example, ``\emph{Allan Donald}'' and ``\emph{The 48-year-old former Test paceman}'' are co-referring mentions.
\textsc{Trans-ShareRepr} allows these tokens to only attend to each other when learning representations using the attention head $h$. 
These tokens are likely to yield similar representations. 
The method thus accomplishes a similar goal as \textsc{Trans-Linking} to allow \emph{tokens} of the same PoC to be treated equivalently during summary generation; we explore the selection of attention heads in \S\ref{sec:experiments}.

\begin{table*}
\setlength{\tabcolsep}{5.5pt}
\renewcommand{\arraystretch}{1.1}
\centering
\begin{small}
\begin{tabular}{|l|cccc|cccc|c|c|c|}
\hline
& \multicolumn{4}{c|}{\textbf{Heuristic Set}} & \multicolumn{7}{c|}{\textbf{Point of Correspondence Test Set}}\\[0.1em]
\textbf{System} & \textbf{R-1} & \textbf{R-2} & \textbf{R-L} & \textbf{BLEU} & \textbf{R-1} & \textbf{R-2} & \textbf{R-L} & \textbf{BLEU} & \textbf{B-Score} & \textbf{\#Tkns} & \textbf{\%Fuse}\\
\hline
\hline
Pointer-Generator & 35.8 & 18.2 & 31.8 & 41.9 & 33.7 & 16.3 & 29.3 & 40.3 & 57.3 & 14.3 & 38.7\\
Transformer & 39.6 & 20.9 & 35.3 & 47.2 & 38.8 & 20.0 & 33.8 & 45.8 & 61.3 & 15.1 & 50.7\\
Trans-\textsc{Linking} & \textbf{39.8} & \textbf{21.1} & \textbf{35.3} & \textbf{47.3} & 38.8 & 20.1 & 33.9 & 45.5 & 61.1 & 15.1 & \textbf{55.8}\\
Trans-\textsc{ShareRepr} & 39.4 & 20.9 & 35.2 & 46.9 & \textbf{39.0} & \textbf{20.2} & \textbf{33.9} & \textbf{45.8} & 61.2 & \textbf{15.2} & 46.5\\
\hdashline
Concat-Baseline & 37.2 & 20.0 & 28.7 & 25.0 & 36.1 & 18.6 & 27.8 & 24.6 & 60.4 & 52.0 & 99.7\\
\hline
\end{tabular}
\end{small}
\caption{Results of various sentence fusion systems.
We report the percentage of output sentences that are generated by fusion (\%Fuse) and the average number of tokens per output sentence (\#Tkns). To calculate \%Fuse, we follow the same procedure used by \citet{lebanoff-etal-2020-understanding} -- a generated sentence is regarded as a fusion if it contains at least two non-stopword tokens from each sentence that do not already exist in the other sentence.}
\label{tab:results_fusion} 
\vspace{-0.1in}
\end{table*}

\begin{figure}
\centering
\includegraphics[width=2.4in]{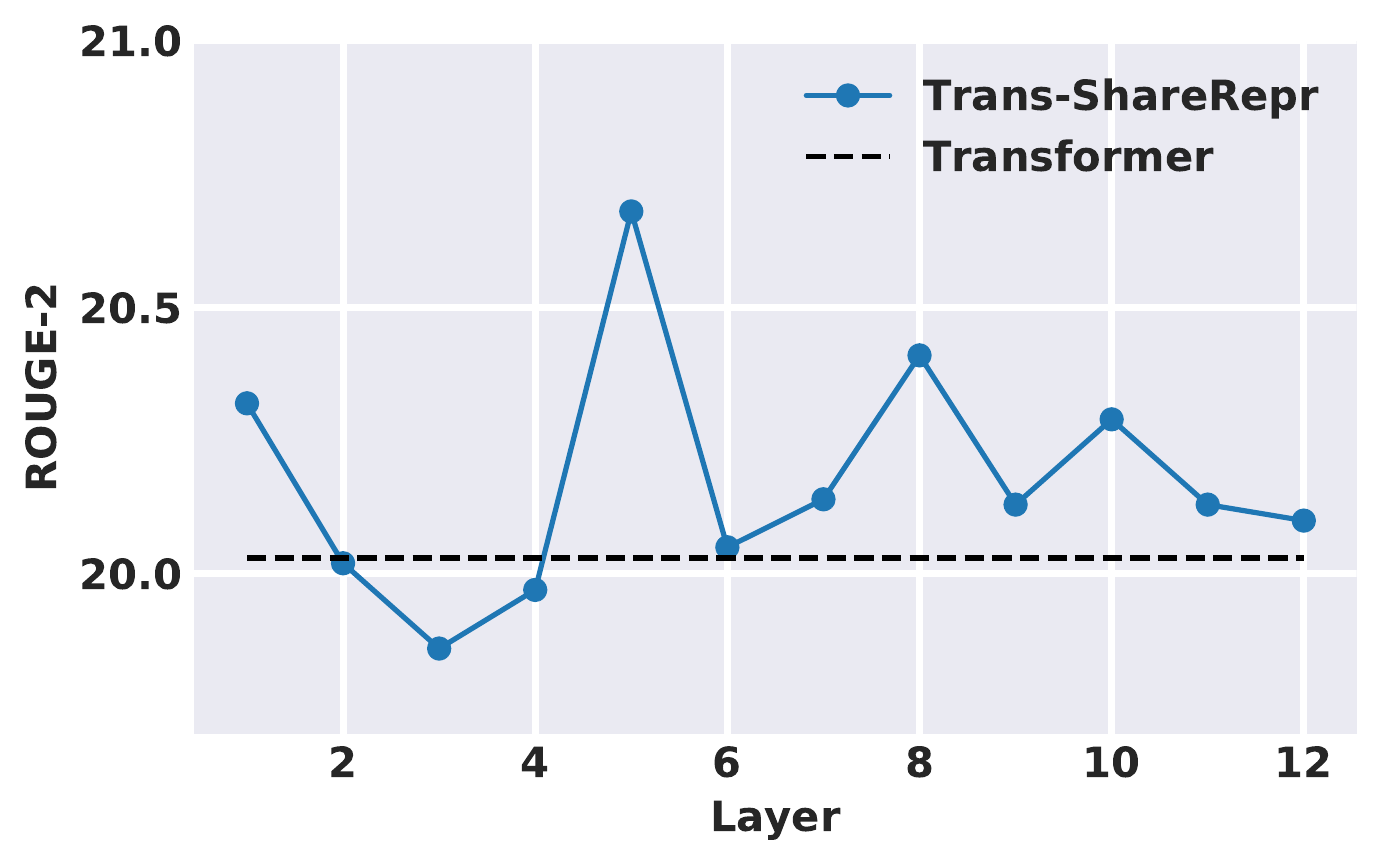}
\vspace{-0.05in}
\caption{The first attention head from the $l$-th layer is dedicated to coreferring mentions.
The head encourages tokens of the same PoC to share similar representations.
Our results suggest that the attention head of the 5-th layer achieves competitive performance, while most heads perform better than the baseline.
The findings are congruent with~\cite{clark-etal-2019-bert} that provides a detailed analysis of BERT's attention.
}
\label{fig:coref_head}
\vspace{-0.1in}
\end{figure}

\section{Experiments}
\label{sec:experiments}

\textbf{Corpus}\quad
Our corpus contains a collection of documents, source and fusion sentences, and human annotations of corresponding regions between sentences. 
The set of documents were sampled from CNN/DM~\cite{see-etal-2017-get} and PoC annotations were obtained from Lebanoff et al.~\shortcite{lebanoff-etal-2020-understanding}.
They use a human summary sentence as an anchor point to find two document sentences that are most similar to it, which forms a fusion instance containing a pair of source sentences and their summary.
PoCs have been annotated based on Halliday and Hasan's theory of cohesion~\shortcite{Halliday:1976} for 1,494 fusion instances, taken from 1,174 documents in the test and valid splits of CNN/DM with a moderate to high inter-annotator agreement (0.58).

\vspace{0.05in}
\textbf{Automatic Evaluation}\quad
We proceed by investigating the effectiveness of various sentence fusion models, including
(a) \textsf{\textbf{\scriptsize Pointer-Generator}}~\cite{see-etal-2017-get} that employs an encoder-decoder architecture to condense input sentences to a vector representation, then decode it into a fusion sentence.
(b) \textsf{\textbf{\scriptsize Transformer}}, our baseline Transformer architecture w/o PoC information.
It is a strong baseline that resembles the UniLM model described in~\cite{NIPS2019_9464}. 
(c) \textsf{\textbf{\scriptsize \textsc{Trans-Linking}}} uses special tokens to mark the boundaries of PoC mentions (\S\ref{sec:linking}).
(d) \textsf{\textbf{\scriptsize \textsc{Trans-ShareRepr}}} allows tokens of the same PoC to share representations (\S\ref{sec:shared_attention}).
All Transformer models are initialized with BERT-\textsc{base} parameters and are fine-tuned using UniLM's sequence-to-sequence objective for 11 epochs, with a batch size of 32.
The source and fusion sentences use BPE tokenization, and the combined input/output sequence is truncated to 128 tokens. 
We use the Adam optimizer with a learning rate of 2e-5 with warm-up. 
For PG, we use the default settings and truncate the output sequences to 60 tokens.

All of the fusion models are trained (or fine-tuned) on the same training set containing 107k fusion instances from the training split of CNN/DM; PoC are identified by the spaCy coreference resolver.
We evaluate fusion models on two test sets, including
a ``heuristic set'' containing testing instances and automatically identified PoC via spaCy, and a final test set containing 1,494 instances with human-labelled PoC.
We evaluate only on the instances that contain at least one point of correspondence, so we have to disregard a small percentage of instances (6.6\%) in the dataset of Lebanoff et al.~\shortcite{lebanoff-etal-2020-understanding} that contain no points of correspondence.

\begin{table}[h]
\setlength{\tabcolsep}{0pt}
\renewcommand{\arraystretch}{1.1}
\centering
\begin{scriptsize}
\textsf{
\begin{tabular}{p{2.9in}}
\hline
\toprule
\textbf{Source:} Later that month, the ICC opened a preliminary examination into the situation in Palestinian territories, \textcolor{red}{paving the way for possible war crimes investigations against Israelis.} \\[0.2em]
Israel and the United States, neither of which is an ICC member, \textcolor{blue}{opposed the Palestinians' efforts to join the body.}\\
\midrule
\textbf{Pointer-Generator:} \emph{ICC opened a preliminary examination into the situation in Palestinian territories .} \\
\midrule
\textbf{Transformer:} \emph{Israel, U.S. and the United States are investigating possible war crimes, \textcolor{red}{paving way for war crimes.}} \\
\midrule
\textbf{Transformer-\textsc{ShareRepr}:} \emph{Israel and U.S. opposed the ICC's investigation into the situation in Palestinian territories.} \\
\midrule
\textbf{Reference:} \emph{Israel and the United States \textcolor{blue}{opposed the move}, which could open the door to war crimes investigations against Israelis.} \\
\bottomrule
\end{tabular}}
\end{scriptsize}
\caption{Example output of sentence fusion systems. 
\textsf{\textbf{\scriptsize PG}} only performs sentence shortening rather than fusion.
\textsf{\textbf{\scriptsize Transformer}} fails to retain the original meaning and \textsf{\textbf{\scriptsize Transformer-\textsc{ShareRepr}}} performs best.
\textsf{\textbf{\scriptsize Reference}} demonstrates a high level of abstraction.
Sentences are manually de-tokenized for readability.}
\label{tab:output}
\end{table}

We compare system outputs and references using a number of automatic evaluation metrics including ROUGE~\cite{lin-2004-rouge}, BLEU~\cite{papineni-etal-2002-bleu} and BERTScore~\cite{Zhang2020BERTScore}. 
Results are presented in Table~\ref{tab:results_fusion}.
We observe that all Transformer models outperform PG, suggesting that these models can benefit substantially from unsupervised pretraining on a large corpus of text.
On the heuristic test set where training and testing conditions match (they both use automatically identified PoC), \textsf{\textbf{\scriptsize \textsc{Trans-Linking}}} performs better than \textsf{\textbf{\scriptsize \textsc{Trans-ShareRepr}}}, and vice versa on the final test set.
We conjecture that this is because the linking model has a stronger requirement on PoC boundaries and the training/testing conditions must match for it to be effective.
In contrast, \textsf{\textbf{\scriptsize \textsc{Trans-ShareRepr}}} is more lenient with mismatched conditions.

We include a \textsf{\textbf{\scriptsize \textsc{Concat-Baseline}}} that creates a fusion by simply concatenating two input sentences. 
Its output contains 52 tokens on average, while other model outputs contain 15 tokens.
This is a 70\% compression rate, which adds to the challenge of content selection~\cite{daume-iii-marcu-2004-generic}.
Despite that all models are trained to fuse sentences, their outputs are not guaranteed to be fusions and shortening of single sentences is possible.
We observe that \textsf{\textbf{\scriptsize \textsc{Trans-Linking}}} has the highest rate of producing fusions (56\%).
In Figure~\ref{fig:coref_head}, we examine the effect of different design choices, where the first attention head of the $l$-th layer is dedicated to PoC. 
We report the averaged results in Table~\ref{tab:results_fusion}.

\vspace{0.05in}
\textbf{Human evaluation}\quad
We investigate the quality of fusions with human evaluation. 
The models we use for comparison include (a) \textsf{\textbf{\scriptsize Pointer-Generator}}, (b) \textsf{\textbf{\scriptsize Transformer}}, (c) \textsf{\textbf{\scriptsize Trans-ShareRepr}} and (d) human reference fusion sentences.
Example outputs for each model can be seen in Table \ref{tab:output}.
We perform evaluation on 200  randomly sampled instances from the point of correspondence test set. 
We take an extra step to ensure all model outputs for selected instances contain fusion sentences, as opposed to shortening of single sentences.
A human evaluator from Amazon Mechanical Turk (\url{mturk.com}) is asked to assess if the fusion sentence has successfully retained the original meaning.
Specifically, an evaluator is tasked with reading the two article sentences and fusion sentence and answering yes or no to the following question, ``Is this summary sentence true to the original article sentences it's been sourced from, and it has not added any new meaning?''
Each instance is judged by five human evaluators and results are shown in Table~\ref{tab:results_human}.
Additionally, we measure their \emph{extractiveness} by reporting on the percentage of $n$-grams ($n$=1/2/3) that appear in the source.
Human sentence fusions are highly abstractive, and as the gold standard, we wish to emulate this level of abstraction in automatic summarizers. Fusing two sentences together coherently requires connective phrases and sometimes requires rephrasing parts of sentences. However, higher abstraction does not mean higher quality fusions, especially in neural models.

\begin{table}
\setlength{\tabcolsep}{2.5pt}
\renewcommand{\arraystretch}{1.1}
\centering
\begin{small}
\begin{tabular}{|l|c|ccc|}
\hline
& & \multicolumn{3}{c|}{\textbf{Extractiveness}}\\
\textbf{System} & \textbf{Truthful.} & \textbf{1-gram} & \textbf{2-gram} & \textbf{3-gram}\\
\hline
\hline
Pointer-Generator & 63.6 & 97.5 & 83.1 & 72.8\\
Transformer & 71.7 & 91.9 & 68.6 & 54.2\\
Trans-\textsc{ShareRepr} & 70.9 & 92.0 & 70.1 & 56.4\\
Reference & 67.2 & 72.0 & 34.9 & 20.9\\
\hline
\end{tabular}
\end{small}
\caption{Fusion sentences are evaluated by their level of truthfulness and extractivenss.
Our system fusions attain a high level of truthfulness with moderate extractivenss.
}
\label{tab:results_human} 
\vspace{-0.1in}
\end{table}

Interestingly, we observe that humans do not always rate reference fusions as truthful.
This is in part because reference fusions exhibit a high level of abstraction and they occasionally contain content not in the source.
If fusion sentences are less extractive, humans sometimes perceive that as less truthful, especially when compared to fusions that reuse the source text.
Our results call for a reexamination of sentence fusion using better evaluation metrics including semantics and question-answering-based metrics~\cite{zhao-etal-2019-moverscore,wang2020asking,durmus-etal-2020-feqa}.

\section{Conclusion}
\label{sec:conclusion}

We address the challenge of information fusion in the context of neural abstractive summarization by making crucial use of points of correspondence between sentences.
We enrich Transformers with PoC information and report model performance on a new test bed for information fusion.
Our findings suggest that modeling points of correspondence is crucial for effective sentence fusion, and sentence fusion remains a challenging direction of research.
Future work may explore the use of points of correspondence and sentence fusion in the standard setting of document summarization.
Performing sentence fusion accurately and succinctly is especially important for summarizing long documents and book chapters~\cite{ladhak-etal-2020-exploring}. These domains may contain more entities and events to potentially confuse a summarizer, making our method of explicitly marking these entities beneficial.

\section*{Acknowledgments}

We are grateful to the anonymous reviewers for their helpful comments and suggestions.
This research was supported in part by the National Science Foundation grant IIS-1909603.

\bibliographystyle{acl_natbib}
\bibliography{logan,anthology}

\end{document}